\documentclass[final,3p,12pt]{elsarticle}
\pdfoutput=1
\usepackage[utf8]{inputenc}
\usepackage[namelimits]{amsmath}
\usepackage{bm}

\usepackage{mathrsfs}

\usepackage[ruled]{algorithm2e}
\usepackage{booktabs}
\usepackage{caption}
\usepackage[colorlinks=true,citecolor=blue,linkcolor=blue]{hyperref}
\usepackage{subfigure}

\usepackage{amssymb}

\usepackage{soul}
\usepackage{color}

\begin{document}

\begin{frontmatter}

\title{Governing equation discovery of a complex system from snapshots}

\author{Qunxi Zhu$^{1,}$\corref{cor1}}
\author{Bolin Zhao$^{2}$}
\author{Jingdong Zhang$^{2}$}
\author{Peiyang Li$^{2}$}
\author{Wei Lin$^{1,2,3}$\corref{cor1}}

\address{$^{1}$Research Institute of Intelligent Complex Systems, Fudan University, Shanghai 200433, China.}

\address{$^{2}$School of Mathematical Sciences, SCMS, SCAM, and CCSB, Fudan University, Shanghai 200433, China.}

\address{$^{3}$Shanghai Artificial Intelligence Laboratory, Shanghai 200232, China.}

\cortext[cor1]{Corresponding authors. E-mails: qxzhu@fudan.edu.cn; wlin@fudan.edu.cn}

\begin{abstract} 
Complex systems in physics, chemistry, and biology that evolve over time with inherent randomness are typically described by stochastic differential equations (SDEs). A fundamental challenge in science and engineering is to determine the governing equations of a complex system from snapshot data. Traditional equation discovery methods often rely on stringent assumptions, such as the availability of the trajectory information or time-series data, and the presumption that the underlying system is deterministic. In this work, we introduce a data-driven, simulation-free framework, called Sparse Identification of Differential Equations from Snapshots (SpIDES), that discovers the governing equations of a complex system from snapshots by utilizing the advanced machine learning techniques to perform three essential steps: probability flow reconstruction, probability density estimation, and Bayesian sparse identification. We validate the effectiveness and robustness of SpIDES by successfully identifying the governing equation of an over-damped Langevin system confined within two potential wells. By extracting interpretable drift and diffusion terms from the SDEs, our framework provides deeper insights into system dynamics, enhances predictive accuracy, and facilitates more effective strategies for managing and simulating stochastic systems.
\end{abstract}

\end{frontmatter}

{\color{black}
The dynamical behaviors of complex systems, such as power grids~\cite{rohden2012self}, ecosystems~\cite{meng2022fundamental}, cellular rhythms~\cite{garcia2004modeling}, bird flocking~\cite{vicsek2012collective}, and climate systems~\cite{marwan2015complex}, are not only captivating but also critical for progress across multiple scientific disciplines. Understanding these behaviors holds the potential to drive significant advances in medical treatments, environmental conservation, and technological innovation~\cite{gao2024learning}. The primary challenges in analyzing complex systems lie in their nonlinearity and stochasticity. Nonlinearity, where small changes in input produce disproportionately large effects on output, is central to the emergence of complex phenomena~\cite{santillan2008use}. This characteristic complicates the modeling and prediction of system behavior, rendering traditional linear approaches inadequate. Stochasticity, which enables a system to adapt to rapidly changing environments, is another fundamental feature of complex systems~\cite{mao2002environmental,zhang2024machine}. Despite the insights offered by experimental observations, uncovering the underlying mechanisms of these nonlinear stochastic dynamics from observational data remains an unresolved challenge and an active area of research. 

To unlock the critical dynamical properties of complex systems, there is an urgent need for efficient approaches to model and interpret stochastic dynamical systems. A key challenge is reconstructing meaningful mathematical models from observational data that accurately represent the underlying complex processes. One promising methodology involves using snapshots of system behavior over time to model stochastic dynamics~\cite{neklyudov2023action}. By extracting continuous underlying dynamics from these snapshots, researchers can capture essential features and statistical properties, facilitating accurate predictions of future behavior and regulation of the system. This approach allows scientists to identify key patterns and behaviors that may not be evident from isolated observations, offering a more comprehensive understanding of system dynamics. These techniques have proven effective in various fields, such as ecology, where temporal snapshots model population dynamics, and finance, where historical data inform market behavior models~\cite{kantz2004nonlinear,posfai2016network}. Leveraging such methodologies enhances our ability to decode the complexities of stochastic systems, advancing research across multiple scientific domains~\cite{vazquez2007porous,schiebinger2019optimal,noe2020machine,tong2020trajectorynet}.

A fundamental requirement for decoding complex systems is the development of a mathematical model that captures the system's evolution over time. Uncertainty is an intrinsic aspect of modeling real-world systems and is widely acknowledged within the scientific community. Conventional modeling algorithms typically assume that uncertainty in the governing equations of complex systems manifests as an additive stochastic perturbation~\cite{course2023state}. In this study, we focus specifically on the stochastic dynamics of complex systems as represented by I$\hat{\text{t}}$o-type stochastic differential equations (SDEs)~\cite{oksendal2013stochastic}. Our objective is to learn SDEs such that the simulated particle densities align with the observed snapshots, providing a more accurate representation of the underlying system dynamics. 

Recent advances in generative modeling have focused on learning stochastic dynamics that interpolate between the data distribution and a prior distribution. Specifically, score-based diffusion models~\cite{songscore,ho2020denoising} utilize an SDE to transition samples from the data distribution to a prior distribution. They employ score matching~\cite{hyvarinen2005estimation} to learn a reverse SDE that captures the gradients of intermediate distributions. However, these methods typically depend on analytical forms of the SDEs and the tractability of intermediate Gaussian distributions.
Additionally, continuous normalizing flows~\cite{chen2018neural,grathwohl2018ffjord}, which are based on Neural ODE methods, have been used in generative models. Yet, during training, simulating the distribution path by numerically integrating the parametrized ODEs at each iteration can lead to prohibitively high computational costs. Recently,  a scalable and simulation-free approach named Flow Matching (FM) is introduced to train probability flows by directly regressing vector fields along specific conditional probability paths~\cite{lipmanflow,zhuswitched}. It is worth noting that two concurrent studies, the stochastic interpolant by~\cite{albergo2022building} and the rectified flow by~\cite{liuflow}, propose similar methodologies for matching distributions using flows, albeit from different perspectives. Notably, the primary difference between our task and mainstream dynamics inference tasks is the availability of temporal trajectory data. In traditional dynamics inference, the evolutionary rules of the system are directly encoded in the temporal trajectory data. In contrast, our approach relies on snapshots, which capture how the dynamical flow transforms the distribution in a more indirect manner.

To enhance the interpretability, sparse identification of nonlinear dynamics plays a critical role in extracting representative expressions from neural network-based models. This technique enables researchers to distill complex neural network models into simpler and interpretable mathematical formulations, facilitating the understanding of complex systems~\cite{brunton2016discovering}. Initiated from a set of basis functions, sparse identification aims at reducing the complexity of the model while preserving essential dynamics with a sparse combination of the basis functions, which is particularly valuable in high-dimensional settings often encountered in neural networks~\cite{gao2022autonomous,course2023state}. In order to obtain the sparsity of the coefficients, the Bayesian inference method based on the sparse prior of the coefficients has been extensively employed~\cite{carvalho2009handling,tipping2001sparse,louizos2017bayesian}.  As the field of machine learning continues to evolve, integrating sparse identification techniques with neural networks may significantly advance our ability to model complex phenomena in various applications, from fluid dynamics to biological systems~\cite{wang2016data}.

We introduce a novel learning method, Sparse Identification of Differential Equations from Snapshots (SpIDES), designed to efficiently uncover the underlying stochastic dynamics of complex systems from snapshot data and to infer the symbolic representations of their models. Our approach utilizes the flow matching and the score matching techniques to learn the probability flow and estimate the score function, respectively, and then integrates these with Bayesian sparse identification based on a library of symbolic differential equations. We demonstrate the effectiveness of SpIDES using a bistable model driven by a double-well potential. Compared to conventional simulation-based machine learning models, our framework simultaneously identifies the distribution transition patterns and discovers interpretable stochastic dynamics without relying on simulations. This leads to reduced computational complexity, enabling scalability to high-dimensional systems.
 
 }

\section{Results}
\subsection{Problem statement}
      To mathematically define our problem statement, consider the process of a complex system governed by an It{\^o} SDE:
      \begin{equation} \label{eq:govern}
      	{\rm d}{\bf x}(t) = {\bf F}({\bf x}, t){\rm d} t + {\bf G}({\bf x}, t) {\rm d} {\bf \beta}(t),
      \end{equation}
      where ${\bf x}(t)=[{x}_1(t), {x}_2(t), ..., {x}_d(t)]^{\top} \in \mathbb{R}^{d}$ represents the state of a system at time $t$, ${\bf F}:\mathbb{R}^d\times \mathbb{R} \rightarrow \mathbb{R}^d$ is the drift function, ${\bf G}:\mathbb{R}^d\times\mathbb{R} \rightarrow \mathbb{R}^{d\times d}$ is the diffusion matrix, and ${\bf \beta}(t): \mathbb{R} \rightarrow \mathbb{R}^{d}$ is the standard Wiener process (a.k.a., Brownian motion). 
      
      Given the snapshot dataset, $\mathcal{D} = \{\mathcal{D}_i\}_{i=1}^{N}$ with $N$ snapshots $\mathcal{D}_i= \{{\bf x}^{(j)}(t_i)\}_{j=1}^{N_i}$, in which $t_i \in [0, T]$ is the timestamp associated with $N_i$ observations ${\bf x}^{(j)}(t_i)\in \mathbb{R}^{d}$. Our goal is to infer the governing equation~\eqref{eq:govern}.
      
      In particular, we leverage the fact that most complex systems have only a few relevant terms that define the dynamics, making the governing equations sparse in a high-dimensional nonlinear function space. To be concrete, our assumption is that the drift function can be approximated by a sparse, linear combination of known basis functions:
      \begin{equation}\label{eq:sparse_drift}
      	{\bf F}({\bf x}, t) = {\bf F}_{\emptyset}({\bf x}, t) + {\bf \theta}_{\text{drift}} \phi_{\text{drift}}({\bf x}, t):={\bf F}_{{\bf \theta}_{\text{drift}}}({\bf x}, t),
      \end{equation}
      where ${\bf F}_{\emptyset}:\mathbb{R}^d\times \mathbb{R} \rightarrow \mathbb{R}^d$ indicates the drift funtion of the known dynamics and $\phi_{\text{drift}}: \mathbb{R}^d\times \mathbb{R} \rightarrow \mathbb{R}^m$ is a library consisting of $M$ candidate nonlinear functions with the coefficient matrix ${\bf \theta}_{\text{drift}} \in \mathbb{R}^{d\times m}$ to be determined. Typically, $\phi_{\text{drift}}({\bf x}, t)$ may include constant, polynomial, and trigonometric terms:
      \begin{equation}
      	\begin{aligned}
      		\phi_{\text{drift}}({\bf x}, t)=[&1, x_1, ..., x_d, t, x_1^2, x_1x_2 ..., t^2,..., \\
      		&\sin(x_1), ..., \sin(x_d), \sin(t),\\
      		&\cos(x_1), ..., \cos(x_d), \cos(t),...]^{\top}.
      	\end{aligned}
      \end{equation}
      Similarly, we write the diffusion matrix as 
      \begin{equation}\label{eq:sparse_diffusion}
      	{\bf G}({\bf x},t) = {\bf G}_{\emptyset}({\bf x},t) + \text{diag}[{\bf \theta}_{\text{diffusion}} \phi_{\text{diffusion}}({\bf x},t)]:={\bf G}_{{\bf \theta}_{\text{diffusion}}}({\bf x},t),
      \end{equation}
      where  ${\bf G}_{\emptyset}: \mathbb{R}^d\times\mathbb{R} \rightarrow \mathbb{R}^{d\times d}$ indicates the known term in the diffusion matrix and $\phi_{\text{diffusion}}: \mathbb{R}^d\times\mathbb{R} \rightarrow \mathbb{R}^n$ is a library consisting of $n$ candidate nonlinear functions with the coefficient matrix ${\bf \theta}_{\text{diffusion}} \in \mathbb{R}^{d\times n}$ to be determined. For simplicity of notation's usage, we denote ${\bf \theta}:=\text{vec}({\bf \theta}_{\text{drift}},{\bf \theta}_{\text{diffusion}})\in \mathbb{R}^{d(m+n)}$, where the operator $\text{vec}(\cdot)$ maps a matrix into a vector by stacking columns. 
      
      In alignment with the assumption of a sparse linear combination of basis functions, we will make use of a sparsity-inducing horseshoe prior over the parameters, ${\bf \theta}\sim p({\bf \theta})$.

      \subsection{Overview of SpIDES}
      We tackle the problem of the sparse identification of differential equations from snapshots using the integrated machine learning techniques, consisting of three important steps: probability flow reconstruction,  probability density estimation, and Bayesian sparse identification. Here, a high-level summary of our framework is provided in Fig. 1 and we introduce the core ingredients of our methodology in the following.
      
      For the SDE~\eqref{eq:govern}, there exists a corresponding deterministic process whose trajectories share the same marginal probability densities $\{p_t({\bf x})\}_{t=0}^{T}$, resulting in the probability flow ODE:
      \begin{equation} \label{eq:PF_ODE}
      	\begin{aligned}
      		{\rm d}{\bf x}(t) &= \left\{
        {\bf F}({\bf x}, t) 
        - \frac{1}{2} \nabla\cdot \left[{\bf G}({\bf x},t){\bf G}({\bf x},t)^{\top}\right]
        - \frac{1}{2} {\bf G}({\bf x},t){\bf G}({\bf x},t)^{\top} \nabla_{{\bf x}} \log p_t({\bf x})\right\}{\rm d} t:={\bf f}({\bf x}, t) {\rm d} t.
      	\end{aligned}
      \end{equation}
      Here, we aim to develop a simulation-free way to approximate the vector field ${\bf f}({\bf x}, t)$ using a neural network, denoted as ${\bf f}_{{\bf \phi}}({\bf x}, t)$ with the trainable weights ${\bf \phi}$. However, the solely available data is the snapshot data $\mathcal{D} = \{\mathcal{D}_i\}_{i=1}^{N}$. We thereby utilize the Flow Matching (FM) algorithm~\cite{lipmanflow,tong2023conditional} to train the neural network. Specifically, we routinely employ the optimal transport to reconstruct trajectories from the snapshot data, and then use the interpolation method to numerically approximate the derivatives $\dot{{\bf x}}^{(j)}(t_i)$ from the trajectories. Notably, in such a way, there is no need to run an ODE solver to numerically obtain the solution of the ODE~\eqref{eq:PF_ODE}, which can significantly reduce the computational complexity, especially for high-dimensional systems. Thus, we can approximate ${\bf f}({\bf x}, t)$ via minimizing the following FM objective:
      \begin{equation}
      	\mathcal{L}_{\text{FM}}({\bf \phi}) = \mathbb{E}_{t, p_t({\bf x})} \| \dot{{\bf x}}(t) - {\bf f}_{{\bf \phi}}({\bf x}, t)\|^2,
      \end{equation}
      where $t$ is sampled from the observed snapshot times, assuming $t_i$, and ${\bf x}(t=t_i)$ is sampled from the corresponding snapshot $\mathcal{D}_{i}$ associated with the preprocessed derivative $\dot{{\bf x}}(t_i)$. 
      
      We further estimate the score function $\nabla_{{\bf x}} \log p_t({\bf x})$ in Eq.~\eqref{eq:PF_ODE} in a simulation-free way based on the discrete-time snapshots $\{\mathcal{D}_{i}\}_{i=1}^N$, utilizing Score Matching (SM) algorithm~\cite{hyvarinen2005estimation} (Fig.1 and Method). The generated score function using a neural network is denoted as ${\bf s}_t({\bf x}; \xi)$, which can be used to approximate the true score function by minimizing the following SM objective:\
      \begin{equation} 
      \mathcal{L}_{\text{SM}}(\xi) = \mathbb{E}_{t,p(\mathbf{x})} \left[ {\rm{tr}}(\nabla_{\bf x} {\bf s}_t({\bf x}; \xi)) + \frac{1}{2}\|{\bf s}_t({\bf x}; \xi)\|^2_2 \right].
    \end{equation} 
      
      With the approximated vector field ${\bf f}_{{\bf \phi}}({\bf x}, t)$ and   the estimated score function ${\bf s}_t({\bf x}; \xi)$, the  vector filed of the probability flow ODE~\eqref{eq:PF_ODE} has the following form when we replace ${\bf f}({\bf x}, t)$ and $\nabla_{{\bf x}} \log p_t({\bf x})$ with ${\bf f}_{{\bf \phi}}({\bf x}, t)$ and ${\bf s}_t({\bf x}; \xi)$, respectively:
      \begin{equation} \label{eq:PF_ODE_noise}
      		{\bf F}({\bf x}, t) 
        - \frac{1}{2} \nabla\cdot \left[{\bf G}({\bf x},t){\bf G}({\bf x},t)^{\top}\right]
        - \frac{1}{2} {\bf G}({\bf x},t){\bf G}({\bf x},t)^{\top}{\bf s}_t({\bf x}; \xi) = {\bf f}_{{\bf \phi}}({\bf x}, t).
      \end{equation}
      
      Next, we employ the sparse Bayesian learning to seek a sparse solution of ${\bf \theta}$ to the overdetermined system~\eqref{eq:PF_ODE_noise} by replacing ${\bf F}({\bf x}, t)$ and ${\bf G}(t)$ with their sparse representations~\eqref{eq:sparse_drift} and \eqref{eq:sparse_diffusion}, respectively. Here, we make use of a sparsity-inducing horseshoe prior~\cite{carvalho2009handling} over the parameter vector ${\bf \theta}$, denoted by $p({\bf \theta})$, and approximate its posterior using the log-normal parametrization, denoted by $q_{{\bf \eta}}({\bf \theta})$ in which ${\bf \eta}$ denotes the vector of variational parameters. Then, the objective for the sparsity is 
      \begin{equation}\label{eq:sparse_loss}
      \begin{aligned}
      \mathcal{L}_{\text{sparsity}}({\bf \eta}) = &\mathbb{E}_{t, p_t({\bf x}), q_{{\bf \eta}}({\bf \theta})} \|{\bf r}_{{\bf \phi},{\bf \theta},{\bf \eta},{\bf \xi}}({\bf x}, t)\|^2\\
      &+\lambda_{\text{KL}}D_{\text{KL}}[q_{{\bf \eta}}({\bf \theta})||p({\bf \theta})],
      \end{aligned}
      \end{equation}
      where ${\bf \theta}$ is sampled from $q_{{\bf \eta}}({\bf \theta})$ using the standard reparametrization trick, ${\bf r}_{{\bf \phi},{\bf \theta},{\bf \eta},{\bf \xi}}({\bf x}, t):={\bf F}_{{\bf \theta}_{\text{drift}}}({\bf x}, t) - \frac{1}{2} \nabla\cdot \left[{\bf G}_{{\bf \theta}_{\text{diffusion}}}({\bf x},t){\bf G}_{{\bf \theta}_{\text{diffusion}}}({\bf x},t)^{\top}\right] -\frac{1}{2} {\bf G}_{{\bf \theta}_{\text{diffusion}}}({\bf x},t){\bf G}_{{\bf \theta}_{\text{diffusion}}}({\bf x},t)^{\top} {\bf s}_t({\bf x}; \xi) - {\bf f}_{{\bf \phi}}({\bf x}, t)$ is the residual, $D_{\text{KL}}[q_{{\bf \eta}}({\bf \theta})||p({\bf \theta})]$ indicates the KL divergence between $q_{{\bf \eta}}({\bf \theta})$ and $p({\bf \theta})$, and $\lambda_{\text{KL}}>0$ is a hyperparameter.
      
      Simply put, we aim to minimize the following objective:
      \begin{equation}
      \mathcal{L}({\bf \phi}, {\bf \eta}) = \mathcal{L}_{\text{derivative}}({\bf \phi}) + \lambda_{\text{sparsity}} \mathcal{L}_{\text{sparsity}}({\bf \eta}) 
      \end{equation}
      with a hyperparameter $\lambda_{\text{sparsity}}>0$.
      In Methods, we describe the detailed data preprocess, reparametrization trick, training procedure, and experimental configurations. In the following sections, we provide some examples of SpIDEs applied to problems in the governing equation
discovery.
  \begin{figure}[ht]
        \centering
        \includegraphics[width=1\linewidth]{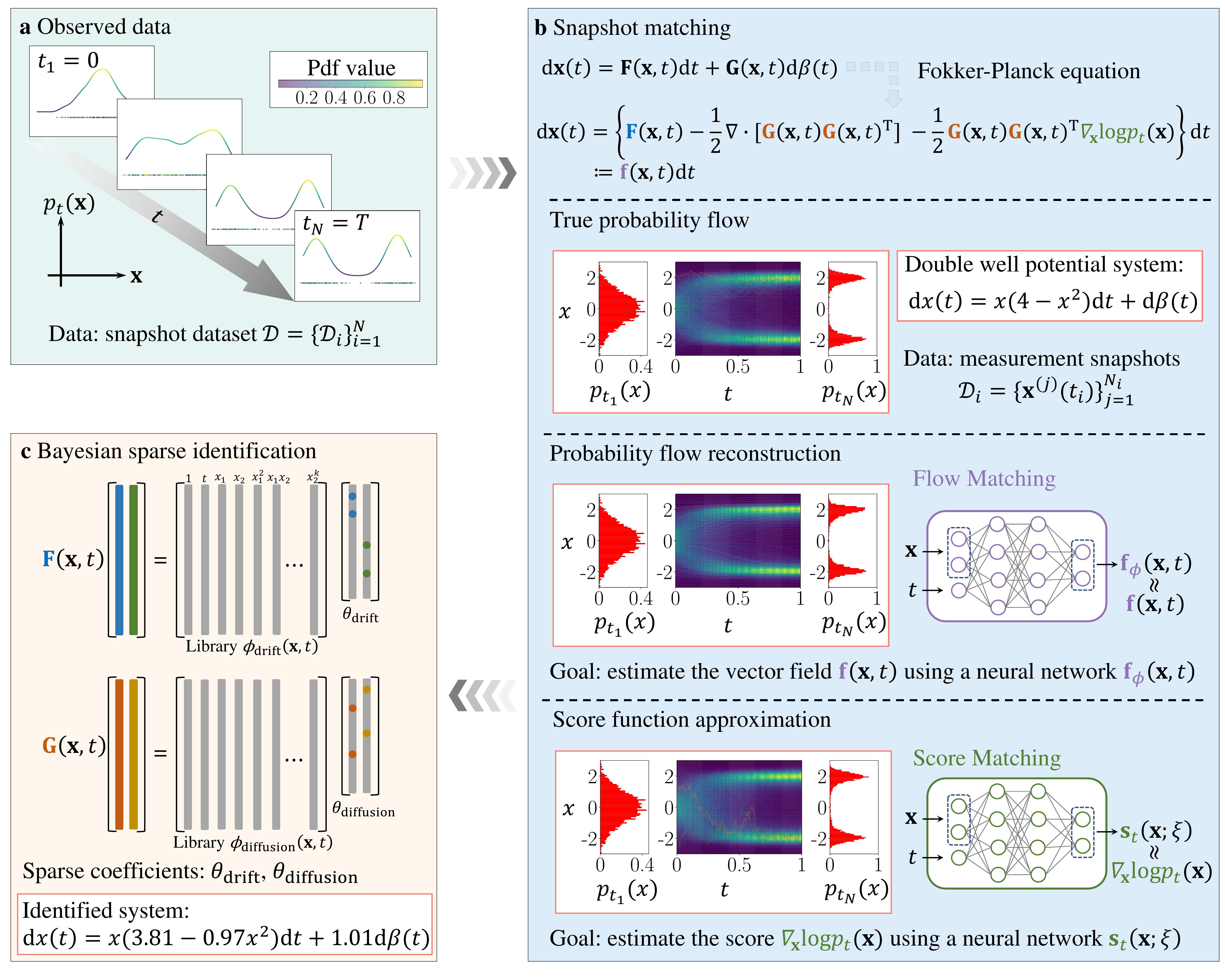}
        \caption{Pipeline of SpIDE: \textbf{a} Observation data include several snapshots at different time steps. \textbf{b} The neural networks predict probability flow and score function through flow matching and score matching respectively. \textbf{c} Bayesian sparse identification with horseshoe prior derives symbolic regression results. }
        \label{1}
        \end{figure}

      \subsection{Example 1: Learning the stochastic dynamics of double potential model}

The double well potential is arguably one of the most important potentials in many fields, such as quantum mechanics and quantum computing \cite{Jelic_2012} In quantum mechanics, the double-well potential function is often used to describe the behavior of electrons within molecules, depicting the motion and distribution of electrons during the interconversion between isomers. In the field of quantum computing, the double-well potential function is a central method for realizing qubits \cite{Realization2016}. Also in the fields of thermodynamic and spectroscopic, there are general approaches for molecules and materials applications through the simulation of quantum vibrational states with double-well potentials \cite{mitoli2024anharmonicvibrationalstatesdoublewell}. The former studies have learned the numerically exact solution of dynamic systems with different potentials \cite{marsiglio_harmonic_2009} and the stochastic analysis of double well potential systems \cite{zhu_stochastic_2013}. Furthermore, it is desirable to learn a more interpretable term of equation through dynamical inference.

To illustrate the process of deducing the dynamic of double well potential, here we conduct simulations of a specific simple version of a stochastic differential equation:
     	\begin{equation} \label{eq:ex1}
      	{\rm d}{x}(t) =x(4-x^2){\rm d} t + {\rm d}{\beta}(t).
      \end{equation}
      Its probability flow ODE is 
      	\begin{equation} \label{eq:ex1_ODE}
      	\dot{x}(t) = x(4-x^2) - \frac{1}{2} \nabla_{{x}} \log {p}_{t}({\bf x}).
      \end{equation}
In the simulations, we uniformly sample snapshots from the interval $[0,1]$ as shown in Fig.\ref{1}a. To infer the underlying double-well potential dynamics, we employ two neural networks: one to reconstruct the vector field corresponding to the probability flow, and the other to estimate the score function. The results in Fig.\ref{1}b display the true probability density path, the probability flow reconstructed by FM, and the probability path generated by the score function derived from SM. A comparison of the three figures clearly shows that FM accurately reconstructs the system's probability flow, while SM provides an accurate approximation of the score function. In Fig.\ref{1}c, we present the outcome of Bayesian sparse identification, showing that the regression coefficients for both drift and diffusion are closely aligned with the true values. This indicates that our method not only successfully selects the correct terms from the library of basis functions but also accurately identifies the corresponding parameters, thus accurately reconstructing the system's governing equation from snapshots.

{\color{black} \section{Discussion}}
In this article, we have introduced the SpIDES framework, which is capable of discovering the governing equation of a complex system from snapshots in a simulation-free manner using advanced machine learning techniques. Specifically, SpIDES consists of three key steps: 1) reconstructing the probability flow via the flow matching method, 2) estimating the score function using the score matching method, and 3) discovering the governing equations using the Bayesian sparse identification method. To conclude, we summarize our findings as follows.

First, mainstream methods for modeling complex systems typically rely on tracking temporal trajectory data to capture the complete underlying dynamics. However, in more realistic scenarios, where only snapshot data is available, these traditional methods may fail. By leveraging flow matching and score matching within a dynamical theory-guided framework, one can efficiently learn the probability flow and estimate the score function, respectively, corresponding to the stochastic dynamics. Accurate reconstruction of the probability flow and estimation of the score function are crucial for the downstream task of discovering governing equations via Bayesian sparse identification.

Second, key properties of real-world complex systems, including many physical systems, are often encoded in concise equations that reflect underlying important structural information, such as symmetries~\cite{liu2022machine,lamb1998time}. Further model distillation and analysis are essential to extract the intrinsic properties of these systems~\cite{zhang2024learning}. Unlike traditional approaches, the SpIDES framework simultaneously performs the sparse identification of both the drift and diffusion terms of the stochastic dynamics with the learned probability flow and estimated score function, thus providing the greater interpretability of the underlying systems.

Looking ahead, several future research directions warrant exploration. First, there is growing interest in higher-order interactions within complex systems~\cite{li2024higher,lambiotte2019networks}. Extending SpIDES to account for interactions between nodes in networked dynamics could significantly advance the field. Additionally, after identifying the stochastic dynamics, one key application is to regulate the system by designing effective control policies~\cite{zhang2022neural,zhangfessnc}.

\section{Methods} 

\subsection{Probability flow reconstruction using (Conditional) Flow Matching}
Recall that, given the SDE~\eqref{eq:govern} and its corresponding deterministic process~\eqref{eq:PF_ODE}, our first goal is to approximate the vector field ${\bf f}({\bf x}, t)$ using a neural network ${\bf f}_{{\bf \phi}}({\bf x}, t)$ based on the snapshot dataset. To achieve this, we employ Flow Matching (FM)~\cite{lipmanflow,tong2023conditional}, a straightforward, simulation-free approach that uses a stable objective by regression based on the target vector field ${\bf f}({\bf x}, t)$. Note that this target vector field generates the desired probability density paths $q_t({\bf x})$, ensuring that $q_{t_i}({\bf x}) = p_{t_i}({\bf x})$ for $i = 1, 2, \dots, N$. The regression objective of FM is 
\begin{equation}
  \label{obj_FM}
  \mathcal{L}_{\text{FM}}({\bf \phi})=\mathbb{E}_{t, p_t({\bf x})} \left\|{\bf f}_{{\bf \phi}}({\bf x}, t) - {\bf f}({\bf x}, t)\right\|^2,
\end{equation}
where $t\sim \mathcal{U}(0, 1)$ and ${\bf x}(t) \sim p_t({\bf x})$. Ideally, as the objective \eqref{obj_FM} approaches zero, the learned vector field ${\bf f}_{{\bf \phi}}({\bf x}, t)$ should generate the target distribution $p_t({\bf x})$. However, in practice, this objective \eqref{obj_FM} is generally computationally intractable due to the lack of explicit closed-forms for both ${\bf f}({\bf x}, t)$ and $p_t({\bf x})$.

To address the computational intractability of the problem, Conditional FM (CFM)~\cite{tong2023conditional} proposes a simple way to construct the target probability path by a mixture of conditional probability paths and introduces a more manageable regression objective that facilitates learning the vector field ${\bf f}_{{\bf \phi}}({\bf x}, t)$. 

Specifically, we incorporate latent variables ${\bf z}(t_i,t_{i+1})$ within any two adjacent time points $t\in [t_i,t_{i+1})$,  which is independent of both ${\bf x}$ and $t$. This allows the marginal probability path to be reformulated as:
\begin{equation} \label{cpd_CFM} 
p_t({\bf x})=\int p_t[{\bf x}|{\bf z}(t_i,t_{i+1})]q[{\bf z}(t_i,t_{i+1})]\rm{d}{\bf z},
\end{equation} 
for any $t\in [t_i,t_{i+1})$, $i=1,2,\cdots, N-1$. Unless otherwise specified, in the following, we consider the case where $t\in [t_k,t_{k+1})$. We denote ${\bf f}[{\bf x}, t|{\bf z}(t_k,t_{k+1})]$ as the conditional vector field that generates the conditional probability path $p_t[{\bf x}|{\bf z}(t_k,t_{k+1})]$, it can be further proven that the marginal vector field obtained by marginalizing the conditional vector field in the following way is precisely the target vector field ${\bf f}({\bf x}, t)$ that generates the desired marginal probability path $p_t({\bf x})$ under some mild conditions, i.e.,
\begin{equation} \label{cvf_CFM} 
{\bf f}({\bf x}, t)=\mathbb{E}_{q[{\bf z}(t_k,t_{k+1})]}\frac{{\bf f}[{\bf x}, t|{\bf z}(t_k,t_{k+1})] p_t[{\bf x}|{\bf z}(t_k,t_{k+1})]}{p_t({\bf x})}.
\end{equation}

Then we can get an unbiased estimator of the marginal vector field through a simpler and tractable regression objective:
\begin{equation} \label{obj_CFM} 
\mathcal{L}_{\text{CFM}}({\bf \phi}) = \sum_{i=1}^{N-1}\mathbb{E}_{t\in [t_i,t_{i+1}), q({\bf z}(t_i,t_{i+1})), p_t[{\bf x}|{\bf z}(t_i,t_{i+1})]} \left| {\bf f}_{{\bf \phi}}({\bf x}, t) - {\bf f}[{\bf x}, t|{\bf z}(t_i,t_{i+1})] \right|^2, 
\end{equation} 
which shares the same gradient with respect to ${\bf \phi}$ as the original FM objective~\eqref{obj_FM}. Typically,  we choose the latent condition ${\bf z}:=({\bf x}_{t_k},{\bf x}_{t_{k+1}})$ and $q[{\bf z}(t_k,t_{k+1})]$ is selected as certain coupling between two distributions $p_{t_k}$ and $p_{t_{k+1}}$, i.e., 
\begin{equation} \label{ex_qz}
q[{\bf z}(t_k,t_{k+1})] := q({\bf x}_{t_k}, {\bf x}_{t_{k+1}}), 
\end{equation}
with ${\bf x}(t)$ being the linear interpolation and the conditionals being Gaussian flows between ${\bf x}_{t_k}$ and ${\bf x}_{t_{k+1}}$ with constant standard deviation $\sigma$: 
\begin{equation} \label{ex_cpd}
p_t[{\bf x}|{\bf z}(t_k,t_{k+1})]=\mathcal{N}\left[{\bf x}\bigg|\frac{t-t_k}{t_{k+1}-t_k}{\bf x}_{t_{k+1}}+\frac{t_{k+1}-t}{t_{k+1}-t_k}{\bf x}_{t_k}, \sigma^2\right], 
\end{equation} 
resulting in a constant velocity vector field conditioned on ${\bf z}$: 
\begin{equation} \label{ex_cvf}
{\bf f}[{\bf x}, t|{\bf z}(t_k,t_{k+1})] = \frac{{\bf x}_{t_{k+1}}-{\bf x}_{t_k}}{t_{k+1}-t_k}. 
\end{equation} 

\subsection{Optimal transport}
When applying the aforementioned CFM, it is necessary to construct the probability distribution of the latent condition $q({\bf z}(t_k,t_{k+1}))$ for $k=1,2,\cdots, N-1$. We therefore employ the optimal transport to couple two adjacent probability distributions $p_{t_k}$ and $p_{t_{k+1}}$.

Optimal transport is a mathematical framework for transforming probability distributions~\cite{villani2009optimal,santambrogio2015optimal,peyre2019computational}. It addresses the problem of finding the most cost-effective way to transport mass from one distribution to another. Formally, given two probability measures $p_{t_k}$ and $p_{t_{k+1}}$, the goal is to determine a transport plan $\pi_{(t_k,t_{k+1})}^*$ that minimizes the transportation cost~\cite{kantorovich2006translocation}:
\begin{equation} 
C(p_{t_k}, p_{t_{k+1}})=\inf_{\pi\in\Pi(p_{t_k}, p_{t_{k+1}})}\int c({\bf x}_{t_k}, {\bf x}_{t_{k+1}}){\rm{d}}\pi({\bf x}_{t_{k}},{\bf x}_{t_{k+1}}), 
\end{equation} 
where $c({\bf x}_{t_k}, {\bf x}_{t_{k+1}})$ represents the cost associated with transporting one unit of mass from ${\bf x}_{t_k}$ to ${\bf x}_{t_{k+1}}$. In this work, we define the cost based on the Euclidean distance, which leads to the squared 2-Wasserstein distance, expressed as:
\begin{equation} 
W(p_{t_k}, p_{t_{k+1}})^2=\inf_{\pi\in\Pi(p_{t_k}, p_{t_{k+1}})}\int ||{\bf x}_{t_{k}}-{\bf x}_{t_{k+1}}||^2{\rm{d}}\pi({\bf x}_{t_{k}},{\bf x}_{t_{k+1}}). 
\end{equation} 

Actually, the squared 2-Wasserstein distance can also be expressed in an equivalent dynamic formulation, referred to as the Benamou-Brenier formula~\cite{benamou1999numerical,brenier2003extended,villani2009optimal}, given by:
\begin{equation} 
W(p_{t_k}, p_{t_{k+1}})^2=\inf_{p_t,{\bf f}}\int_{\mathbb{R}^d} \int_{t_k}^{t_{k+1}} p_t({\bf x})||{\bf f}({\bf x}, t)||^2{\rm{d}}t{\rm{d}}{\bf x}. 
\end{equation} 
with the probability density path $p_t$ and the vector field ${\bf f}$ subject to the continuity equation constraint:
\begin{equation} 
\partial_t p_t + \nabla \cdot (p_t{\bf f}) = 0. 
\end{equation} 

In this work, we sample the latent condition ${\bf z}(t_k,t_{k+1})$ from the 2-Wasserstein optimal transport plan $\pi^*_{(t_k,t_{k+1})}$ with marginals $p_{t_k}$ and $p_{t_{k+1}}$, i.e.,
\begin{equation} 
q[{\bf z}(t_k,t_{k+1})]=\pi^*_{(t_k,t_{k+1})}({\bf x}_{t_{k}}, {\bf x}_{t_{k+1}}).
\end{equation} 
In the CFM algorithm, we take the same $p_t[{\bf x}|{\bf z}(t_k,t_{k+1})]$ and ${\bf f}[{\bf x}, t|{\bf z}(t_k,t_{k+1})]$ as in \eqref{ex_cpd} and \eqref{ex_cvf} respectively. It can be further proven that, as the standard deviation $\sigma \to 0$, the optimal transport-based CFM is equivalent to dynamic optimal transport.

\subsection{Score function estimation using Score Matching}
As shown in \eqref{eq:govern}, after obtaining a parametric estimate of ${\bf f}({\bf x}, t)$, it is still necessary to estimate the score function $\nabla_{{\bf x}} \log p_t({\bf x})$ in order to further identify the system's governing equations. Here, we utilize Score Matching (SM)~\cite{hyvarinen2005estimation}, which is a widely used method for estimating the score function without the need for explicit knowledge of the probability distribution.

The SM method minimizes the difference between the true score function $\nabla_{\mathbf{x}} \log p_t({\bf x})$ and a model approximation ${\bf s}_t({\bf x}; \xi)$ parameterized using a neural network. The objective function for score matching, based on minimizing the Fisher divergence, is given by:
\begin{equation} 
\mathcal{L}_{\text{SM}}(\xi) = \mathbb{E}_{t,p(\mathbf{x})} \left[ \|{\bf s}_t({\bf x}; \xi) - \nabla_{\mathbf{x}} \log p_t({\bf x})\|^2 \right].
\end{equation} 

Evidently, directly minimizing this divergence remains challenging, as the objective depends on the unknown target probability density path $p_t({\bf x})$. However, a key insight in score matching is that we can get an equivalent formulation of the above regression objective up to a constant, which depends solely on the model score function:
\begin{equation} 
\hat{\mathcal{L}}_{\text{SM}}(\xi) = \mathbb{E}_{t,p(\mathbf{x})} \left[ {\rm{tr}}(\nabla_{\bf x} {\bf s}_t({\bf x}; \xi)) + \frac{1}{2}\|{\bf s}_t({\bf x}; \xi)\|^2_2 \right].
\end{equation} 

\subsection{Bayesian sparse identification of governing equations}
After estimating the vector field ${\bf f}({\bf x}, t)$ and the score function $\nabla_{\mathbf{x}} \log p_t({\bf x})$, we aim to further identify the underlying governing equations of the data through sparse identification. Specifically, we substitute the sparse representation of the drift and diffusion terms (Eq.~\eqref{eq:sparse_drift} and Eq.~\eqref{eq:sparse_diffusion}) on a library of basis functions into Eq.~\eqref{eq:PF_ODE_noise}, resulting in an overdetermined regression problem:
\begin{equation} 
\begin{aligned}
& {\bf F}_{{\bf \theta}_{\text{drift}}}({\bf x}, t) - \frac{1}{2} \nabla\cdot \left[{\bf G}_{{\bf \theta}_{\text{diffusion}}}({\bf x},t){\bf G}_{{\bf \theta}_{\text{diffusion}}}({\bf x},t)^{\top}\right] -\frac{1}{2} {\bf G}_{{\bf \theta}_{\text{diffusion}}}({\bf x},t){\bf G}_{{\bf \theta}_{\text{diffusion}}}({\bf x},t)^{\top} {\bf s}_t({\bf x}; \xi) \\
= ~&{\bf f}_{{\bf \phi}}({\bf x}, t).
\end{aligned}
\end{equation} 
We utilize a Bayesian approach based on sparse priors to solve this problem.

To promote sparsity in the linear combination parameters ${\bf \theta}$ associated with the basis functions, we assign each parameter $\theta_i$ a horseshoe prior~\cite{carvalho2009handling,carvalho2010horseshoe,course2023state}, denoted as $p(\theta_i)$. Specifically, the horseshoe prior assumes that the parameters $\theta_i$ are conditionally independent with a probability density function $p(\theta_i|\tau)$, which can be expressed as a scale mixture of normal distributions:
\begin{equation} 
(\theta_i|\lambda_i, \tau)\sim \mathcal{N}(0,\lambda_i^2 \tau^2),
\end{equation}
\begin{equation}
\lambda_i \sim \text{C}^{+}(0,1),
\end{equation}
\begin{equation}
\tau \sim \text{C}^{+}(0,\tau_0),
\end{equation}
where $\text{C}^{+}(0,s)=2\{s\pi[1+(z/s)^2]\}^{-1}$ denotes a half-Cauchy distribution, $\lambda_i$ is the local shrinkage parameter, and $\tau$ is the global shrinkage parameter with free parameter $\tau_0$ which can be tuned for specific desiderata. 

Rather than working directly with the horseshoe priors, we will utilize a decomposition based on (inverse) gamma distributions for ease of computation~\cite{neville2014mean,louizos2017bayesian}.  Specifically, the horseshoe prior can be expressed as $\theta_i=\Tilde{\theta_i}\sqrt{s_a s_b \alpha_i \beta_i}$, where
\begin{equation} 
\Tilde{\theta_i}\sim\mathcal{N}(0,1), s_a\sim \mathcal{G}(0.5,\tau_0^2), s_b\sim \mathcal{IG}(0.5,1), \alpha_i\sim \mathcal{G}(0.5,1), \beta_i\sim \mathcal{IG}(0.5,1),
\end{equation}
$\mathcal{G}$ and $\mathcal{IG}$ represent the Gamma and inverse Gamma distributions,respectively.

After specifying the prior, we also need to define an approximate posterior for the parameters $\bf \theta$. It is worth noting that the improper log-uniform prior arises as a limiting case of the horseshoe prior when the shapes of the (inverse) Gamma hyperpriors on ${\alpha}_i$ and ${\beta}_i$ approach zero~\cite{carvalho2010horseshoe, armagan2011generalized}. Based on the mean-field assumption, we employ a log-normal parametrization over the shrinkage parameters, denoted concisely as $q_{{\bf \eta}}({\bf \theta})$:
\begin{equation} 
q_{{\bf \eta}}({\bf \theta})=q_{{\bf \eta}}(s_a, s_b)\prod_{i=1}^{d(m+n)}q_{{\bf \eta}}(\alpha_i, \beta_i)q_{{\bf \eta}}(\Tilde{\theta_i}),
\end{equation}
where ${\bf \eta}$ represents the vector of variational parameters~\cite{louizos2017bayesian,course2023state}:
\begin{equation} 
q_{{\bf \eta}}(s_a, s_b)=\mathcal{LN}(s_a|\mu_{s_a},\sigma_{s_a}^2)\mathcal{LN}(s_b|\mu_{s_b},\sigma_{s_b}^2),
\end{equation}
\begin{equation} 
q_{{\bf \eta}}(\alpha_i, \beta_i)=\mathcal{LN}(\alpha_i|\mu_{\alpha_i},\sigma_{\alpha_i}^2)\mathcal{LN}(\beta_i|\mu_{\beta_i},\sigma_{\beta_i}^2),
\end{equation}
\begin{equation} 
q_{{\bf \eta}}(\Tilde{\theta_i})=\mathcal{N}(\Tilde{\theta_i}|\mu_{\Tilde{\theta_i}},\sigma_{\Tilde{\theta_i}}^2).
\end{equation}
With the aforementioned prior and posterior distributions, the KL divergence between the approximate posterior and the prior in Eq.~\eqref{eq:sparse_loss} can be factorized as:
\begin{equation} 
\begin{aligned}
D_{\text{KL}}(q_{{\bf \eta}}({\bf \theta})||p({\bf \theta}))= &D_{\text{KL}}(q_{{\bf \eta}}(s_b)||p(s_b))+D_{\text{KL}}(q_{{\bf \eta}}({\bf \alpha})||p({\bf \alpha}))+
D_{\text{KL}}(q_{{\bf \eta}}({\bf \beta})||p({\bf \beta}))\\
&+D_{\text{KL}}(q_{{\bf \eta}}({\bf \Tilde{\theta}})||p({\bf \Tilde{\theta}})).
\end{aligned}
\end{equation}
We can further calculate the closed-form of each term as:
\begin{equation} 
D_{\text{KL}}(q_{{\bf \eta}}(s_b)||p(s_b))=\text{exp}(\frac{1}{2}\sigma_{s_b}^2-\mu_{s_b})-\frac{1}{2}(2\text{log}\sigma_{s_a}-\mu_{s_b}+\text{log}2+1),
\end{equation}
\begin{equation} 
D_{\text{KL}}(q_{\boldsymbol{\eta}}({\bf \alpha})||p({\bf \alpha}))=\sum_{i=1}^{d(m+n)}\left[\text{exp}(\frac{1}{2}\sigma_{\alpha_i}^2-\mu_{\alpha_i})-\frac{1}{2}(2\text{log}\sigma_{\alpha_i}-\mu_{\alpha_i}+\text{log}2+1)\right],
\end{equation}
\begin{equation} 
D_{\text{KL}}(q_{{\bf \eta}}({\bf \beta})||p({\bf \beta}))=\sum_{i=1}^{d(m+n)}\left[\text{exp}(\frac{1}{2}\sigma_{\beta_i}^2-\mu_{\beta_i})-\frac{1}{2}(2\text{log}\sigma_{\beta_i}-\mu_{\beta_i}+\text{log}2+1)\right],
\end{equation}
\begin{equation} 
D_{\text{KL}}(q_{{\bf \eta}}({\bf \Tilde{\theta}})||p({\bf \Tilde{\theta}}))=-\frac{1}{2}\sum_{i=1}^{d(m+n)}(2\text{log}\sigma_{\Tilde{\theta_i}}-\mu_{\Tilde{\theta_i}}^2-\sigma_{\Tilde{\theta_i}}^2+1).
\end{equation}
We further use Adam~\cite{kingma2014adam} to optimize $\mathcal{L}_{\text{sparsity}}({\bf \eta})$ with respect to the variational parameters.

\section*{Acknowledgments}
Q.Z. is supported by the National Natural Science Foundation of China (No. 62406072), by the China Postdoctoral Science Foundation (No. 2022M720817), and by the STCSM (Nos. 21511100200, 22ZR1407300, and 22dz1200502). W.L. is supported by the National Natural Science Foundation of China (No. 11925103) and by the STCSM (Nos. 22JC1402500, 22JC1401402, and 2021SHZDZX0103).

{\color{black}
\section*{ Conflicts of Interest}
The Authors declare no Competing Financial or Non-Financial Interests.
}

\bibliographystyle{unsrt} 
\bibliography{main} 

\end{document}